\documentclass[10pt,letterpaper]{article}

\usepackage{cogsci}

\cogscifinalcopy 

\usepackage{pslatex}
\usepackage{apacite}
\usepackage{float} 

\usepackage{todonotes}

\title{Different kinds of cognitive plausibility: why are transformers better than RNNs at predicting N400 amplitude?}
 
\author{{\large \bf James A. Michaelov (j1michae@ucsd.edu)}\\{\large \bf Megan D. Bardolph (mbardolph@ucsd.edu)}\\{\large \bf Seana Coulson (scoulson@ucsd.edu)}\\{\large \bf Benjamin K. Bergen (bkbergen@ucsd.edu)} \\
Department of Cognitive Science, University of California, San Diego\\
9500 Gilman Dr, La Jolla, CA 92093, USA}

\begin{document}

\maketitle

\begin{abstract}
Despite being designed for performance rather than cognitive plausibility, transformer language models have been found to be better at predicting metrics used to assess human language comprehension than language models with other architectures, such as recurrent neural networks. Based on how well they predict the N400, a neural signal associated with processing difficulty, we propose and provide evidence for one possible explanation---their predictions are affected by the preceding context in a way analogous to the effect of semantic facilitation in humans.

\end{abstract}

\section{Introduction}
Neural language models (NLMs) are valuable tools in understanding human language comprehension because they learn to predict language based on the surface-level statistics of language alone. As such, they are inherently models both of what can be predicted and what can be learned about language based only on linguistic input. For this reason, they can be used to test hypotheses about how such knowledge may be used in the human language comprehension system that would be impossible to test experimentally.

Since the early days of their implementation, recurrent neural network language models (RNN-LMs) have been used to investigate human cognition \cite{elman_1990_FindingStructureTimea}. However, the recent development of the transformer network \cite{vaswani_2017_attention} has largely overshadowed them in the field of machine learning. Research has shown that language model performance (operationalized as perplexity) tends to correlate with how human-like RNN-LMs are in their processing of language (e.g., \citeNP{aurnhammer_2019_EvaluatingInformationtheoreticMeasures}). Thus, the fact that transformer language models (T-LMs) perform better than the previously state-of-the-art RNN-LMs despite their vastly different architectures suggests that they may be a viable alternative model of human language processing.

It is therefore unsurprising that in the last year, researchers interested in the cognition of language have compared RNN-LMs and T-LMs in terms of how well they capture human linguistic behavior \cite{merkx_2020_ComparingTransformersRNNs,ettinger_2020_WhatBERTNot,misra_2020_ExploringBERTSensitivity,wilcox_2020_PredictivePowerNeural,eisape_2020_ClozeDistillationImproves}. To the best of our knowledge, only one (currently unpublished) study has attempted to investigate  whether RNN-LMs or T-LMs are better for predicting the N400, a neural response reflecting semantic retrieval demands. In their study, \citeA{merkx_2020_ComparingTransformersRNNs} find that the surprisal of T-LMs fit the human N400 data better than the surprisal of RNN-LMs. This is somewhat unexpected because, as \citeA{merkx_2020_ComparingTransformersRNNs} note, intuitively, RNN-LMs more closely match what we believe about the human language comprehension system---they process language word-by-word and have a limited `working memory'. 

Here we ask whether the success T-LMs evince in predicting N400 amplitude \cite{merkx_2020_ComparingTransformersRNNs} is connected to other findings; for example, they appear to show semantic priming effects and, like the N400 \cite{nieuwland2008truth}, they are rather insensitive to negation \cite{ettinger_2020_WhatBERTNot,misra_2020_ExploringBERTSensitivity}. To do so, we investigate whether the predictions of T-LMs show an analog of semantic priming phenomena that have been argued to impact N400 amplitude to a greater or lesser extent (e.g., \citeNP{brouwer_2012_GettingRealSemantica,lau_2013_DissociatingN400Effects}). If this is the case, it would suggest that T-LMs may be analogous to the human language comprehension system in a different way to RNN-LMs, and may offer an insight into the relationship between these phenomena. Such a result would also indicate that we need to update our conception of what makes a computational language model more or less cognitively plausible.

\section{Background}
The N400 \cite{kutas_1980_ReadingSenselessSentences} is a negative deflection in the event-related brain potential (ERP), peaking roughly 400ms after the presentation of a stimulus. It is thought to index processing difficulty---if the preceding context activates semantic content associated with an upcoming word, the word is easier to process, and thus elicits a reduced amplitude N400. Recent accounts have hypothesized that the N400 specifically indexes the extent to which the upcoming word was not expected; a prediction error not affected by the strength of failed predictions \cite{vanpetten_2012_PredictionLanguageComprehension,luke_2016_LimitsLexicalPrediction,delong_2020_ComprehendingSurprisingSentences,kuperberg_2020_TaleTwoPositivities}.

The surprisal of an NLM towards a word is a clear conceptual analog of the N400---surprisal is the negative logarithm of the probability of an upcoming word given its context. NLM surprisal significantly predicts N400 amplitude, beating other metrics derived from NLMs \cite{frank_2015_ERPResponseAmount,aurnhammer_2019_EvaluatingInformationtheoreticMeasures,merkx_2020_ComparingTransformersRNNs}. Additionally, on more fine-grained analysis, surprisal appears to behave analogously to N400 amplitude---in many cases, experimental manipulations that impact N400 amplitude affect surprisal values in an analogous fashion \cite{michaelov_2020_HowWellDoes}.

However, while surprisal is a good model of the extent to which a word is predicted in the context of a sentence, N400 amplitude is also modulated by other factors \cite{kutas_2011_ThirtyYearsCounting,kuperberg_2020_TaleTwoPositivities}. One key finding is that the N400 to a target word is less negative in amplitude when it follows a semantically related word than an unrelated one \cite{kutas_1988_EventrelatedBrainPotential,kutas_2011_ThirtyYearsCounting}. Additionally, the N400 response to a word is reduced if it is semantically related to the most predictable upcoming word---for example, the word \textit{monopoly} in \textit{``Checkmate,'' Rosaline announced with glee. She was getting to be really good at monopoly} elicits a less negative N400 than the word \textit{football} by virtue of being more semantically related to the best completion, \textit{chess} \cite{federmeier_1999_RoseAnyOther}. The N400 response to a word is also reduced if it is semantically related to the previous words in the utterance---for example, there is no difference in N400 response between the word \textit{eat} in \textit{for breakfast the boys would only eat...} and \textit{for breakfast the eggs would only eat...}, despite its semantic implausibility in the second clause \cite{kuperberg_2003_ElectrophysiologicalDistinctionsProcessing}.

These effects have led some researchers to argue that the N400 can be explained by spreading activation, where words in the preceding context partially activate semantically-related upcoming words, regardless of whether they are appropriate completions to the sentence \cite{brouwer_2012_GettingRealSemantica}. This `bag-of-words' approach to semantic pre-activation \cite{kuperberg_2016_SeparateStreamsProbabilistic} has also been reflected in computational modeling of the N400. Several researchers have used the cosine distance between the word embeddings of the target word and those for the preceding words to reflect the kind of semantic similarity that may lead to facilitated processing \cite{parviz_2011_UsingLanguageModels,ettinger_2016_ModelingN400Amplitude,frank_2017_WordPredictabilitySemantic}. This approach is also used to control for confounding effects of semantic similarity when explicitly investigating prediction \cite{kuperberg_2020_TaleTwoPositivities}.

The present study has two main aims. First, to directly compare and quantify how well semantic facilitation (as operationalized by cosine distance between embeddings) and prediction (as operationalized by surprisal) predict N400 amplitude. Second, to identify the extent to which the two are correlated and how this varies by language model architecture, with the hope that this will inform why T-LMs have been found to predict N400 amplitude better than RNN-LMs despite their apparent cognitive implausibility.

\section{Experiment 1: RNN-LM vs. T-LM surprisal}
\subsection{Modeling approach and details}
This experiment follows the same general approach as previous research investigating surprisal as a predictor of N400 amplitude \cite{frank_2015_ERPResponseAmount,aurnhammer_2019_EvaluatingInformationtheoreticMeasures,merkx_2020_ComparingTransformersRNNs}, namely, comparing recorded ERP data to NLM surprisal for the same set of stimuli. We use stimuli from an ERP study whose data have been previously presented \cite{bardolph_2018_SingleTrialEEG}. These stimuli were run through two NLMs, one RNN-LM and one T-LM, and the predicted probability of target words was collected. This predicted probability was then transformed into surprisal, where the surprisal $S$ of a word $w_i$ is the negative logarithm of its probability given its preceding context $w_1...w_{i-1}$, as shown in (\ref{eq:surprisal}).

\vspace{-2mm}

\begin{equation}
    S(w_{i}) = -\log P(w_{i}|w_{1}...w_{i-1})
    \label{eq:surprisal}
\end{equation}

This surprisal was then used as a predictor in a linear mixed-effects model to predict by-trial, by-electrode amplitude from the original ERP study.

Two NLMs were used. The RNN-LM was the \textsc{BIG LSTM+CNN Inputs} model \cite{jozefowicz_2016_ExploringLimitsLanguage}, henceforth the JRNN. The T-LM used was GPT-2 \cite{radford_2019_LanguageModelsAre}. Both models are very large---the JRNN has roughly 1 billion parameters, while GPT-2 has roughly 1.5 billion. One area in which the two differ is that the JRNN has a vocabulary size of roughly 800,000, while GPT-2 has a vocabulary size of roughly 50,000. Additionally, while the JRNN was trained on approximately one billion words, GPT-2 was trained on a dataset an order of magnitude larger.

\subsection{Original ERP Study}
The original study \cite{bardolph_2018_SingleTrialEEG} used stimuli adapted from previous work \cite{thornhill_2012_LexicalConceptualAnticipation}. There were 290 sentence frames with target words in one of four conditions, for a total of 1160 sentences. 

The conditions were the following. The \textsc{Best Completion} was the completion with the highest cloze probability (\citeNP{taylor_1957_ClozeReadabilityScores}; cloze = $0.458 \pm 0.261$). The cloze probability of a word is the proportion of participants in a norming study that filled in a gap in the sentence with that word. \textsc{Related} completions were low-cloze ($0.043 \pm 0.058$) words  that are semantically related to the best completion. \textsc{Unrelated} completions are low-cloze ($0.024 \pm 0.037$) words that are semantically unrelated to the best completion. \textsc{Implausible} completions are completions that were semantically implausible with a cloze of zero. The conditions can be illustrated with the following example: \textit{It's hard to admit when one is} \textbf{\textit{wrong}} (\textsc{Best Completion}) / \textbf{\textit{incorrect}} (\textsc{Related} to best completion) / \textbf{\textit{lonely}} (\textsc{Unrelated} to best completion) / \textbf{\textit{screened}} (\textsc{Implausible}).

As expected from previous research, \citeA{bardolph_2018_SingleTrialEEG} found that the \textsc{Best Completion} elicited the lowest-amplitude N400, followed by \textsc{Related}, \textsc{Unrelated}, and \textsc{Implausible} completions, in order of increasing amplitude.

In the study, 44 healthy adult experimental participants read sentences in English one word at a time. EEG was recorded from 29 scalp sites. In the present study, the mean amplitude at each site over the 300-500ms time period (the canonical N400 time-frame) was calculated for each electrode in each trial. These mean amplitude measurements thus served as the outcome measures in the regression models described below.

\subsection{Results}
Linear mixed-effects models were used to predict N400 amplitude. All models included region of interest (ROI; Prefrontal, Fronto-central, Central, Posterior, Left Temporal, Right Temporal) as a fixed effect and Subject, Sentence Frame, and Electrode as random intercepts (more complex random effects structures led to models that did not converge or had singular fits).

\begin{figure}
    \centering
    \includegraphics[width = \columnwidth]{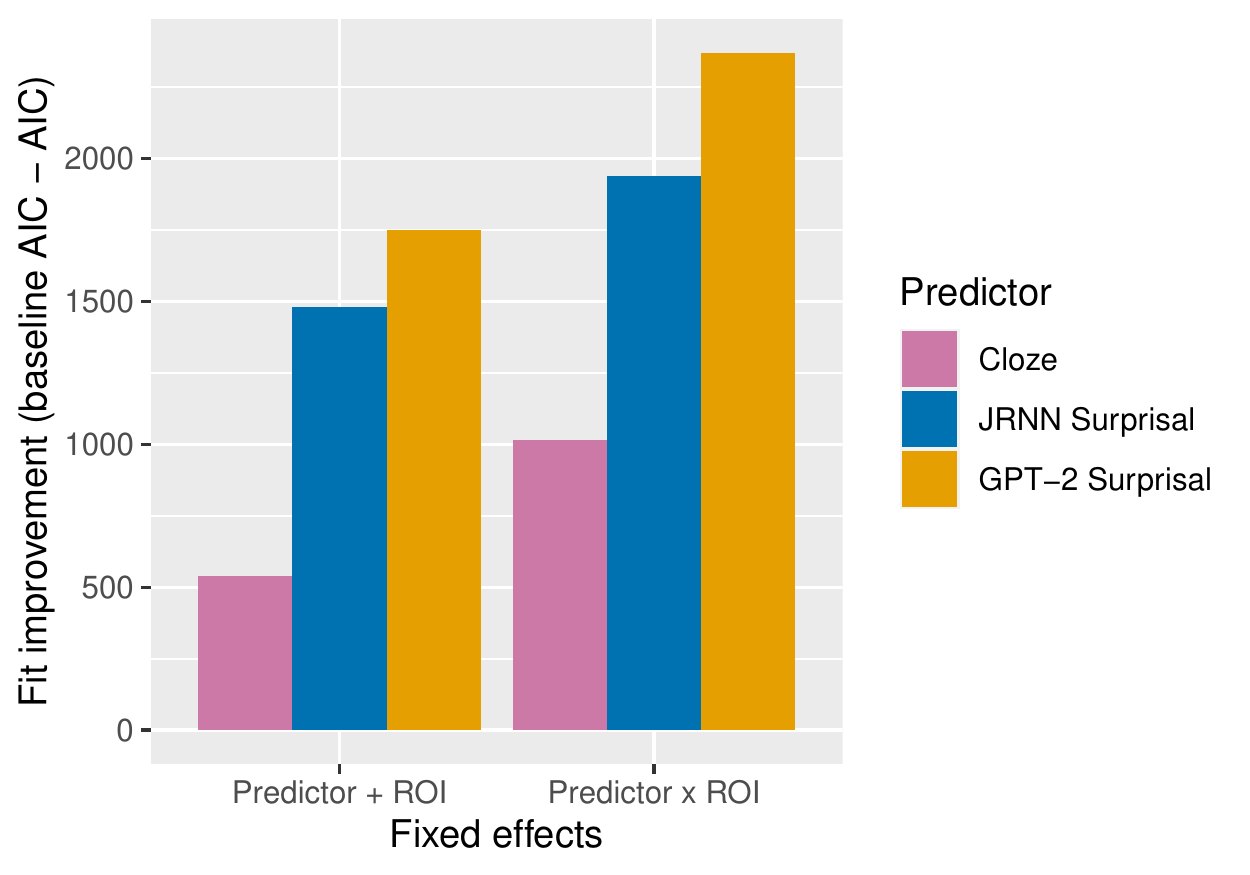}
    \caption{The improvement in AIC for each model compared to the baseline model with only ROI as a main effect. The `Predictor + ROI' models have main effects of the predictor and ROI, while the `Predictor x ROI' model also includes their interaction.}
    \label{fig:AIC}
\end{figure}

We evaluated the statistical significance of each predictor using likelihood ratio tests between nested models.  All reported p-values are corrected for multiple comparisons based on the false discovery rate \cite{benjamini2001control,R_language}. Adding the fixed effect of surprisal to a null model (a linear mixed-effects model with only ROI as a fixed effect and the aforementioned random intercepts) significantly improved model fit (JRNN: $\chi^2(1) = 1483, p < 0.0001$; GPT-2: $\chi^2(1) = 1752.5, p < 0.0001$). Further adding the interaction of surprisal and ROI also significantly improved the model fit (JRNN: $\chi^2(5) = 467.82, p < 0.0001$; GPT-2: $\chi^2(5) = 628.84, p < 0.0001$).

A comparison of AICs (Figure \ref{fig:AIC}) shows that for equivalent models, GPT-2 surprisal fits N400 amplitude more closely. For comparison, we also include the AIC values for equivalent linear mixed-effects models with cloze probability as the main predictor. While it may be unsurprising that cloze is a worse predictor of N400 amplitude than surprisal in this study due to the fact that the  \textsc{Related} and \textsc{Unrelated} conditions were matched for cloze, as far as we are aware, this is the first time that a corpus-derived metric has out-performed cloze as a predictor of human processing difficulty (see \citeNP{brothers_2021_WordPredictabilityEffects}, for discussion). To investigate the difference in AICs further, we compared the predictions of the best models (i.e. those including the Predictor x ROI interaction) to the real N400 in a held-out dataset of roughly 15\% of the total data (46,280 measurements). 

\begin{figure}
    \centering
    \includegraphics[width = \columnwidth]{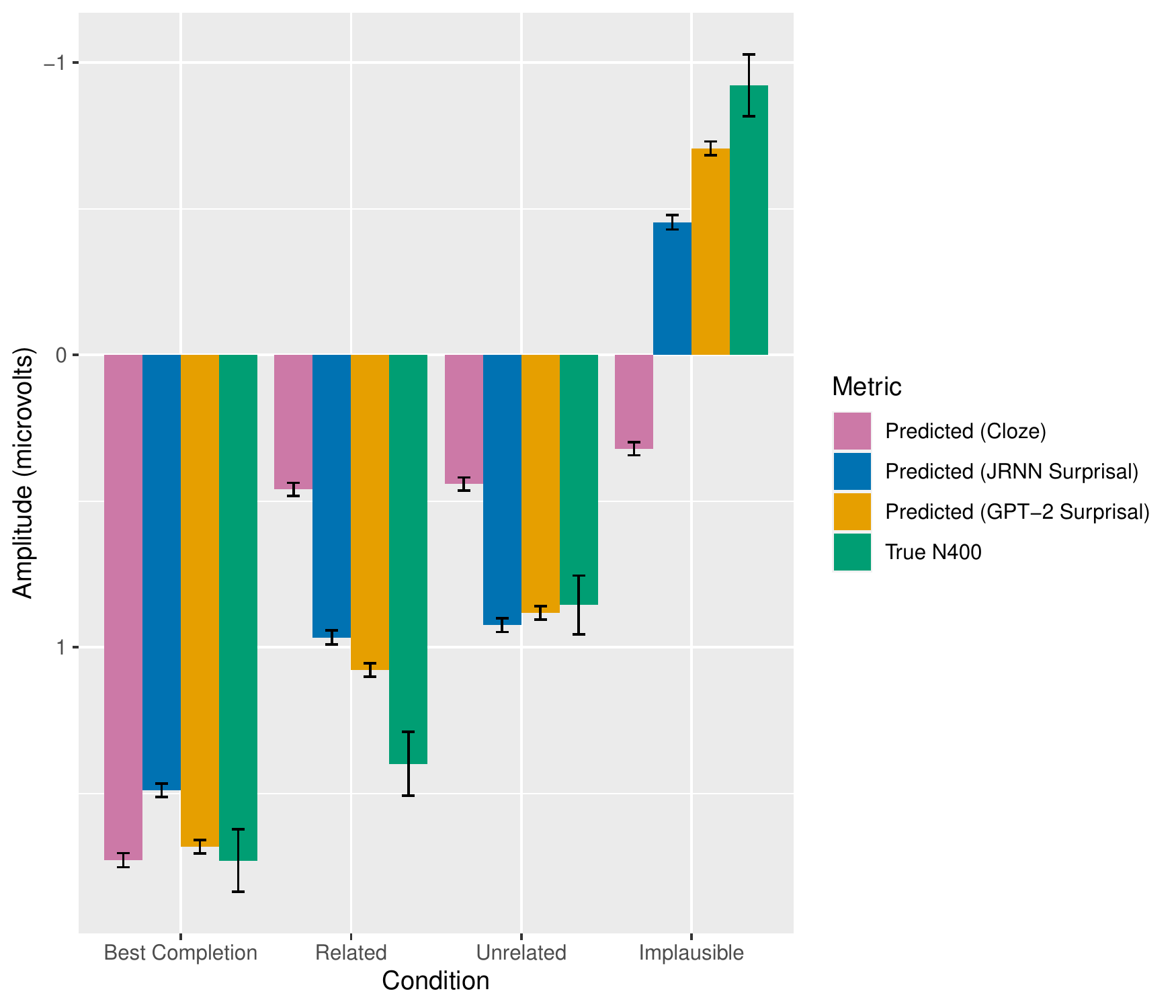}
    \caption{True and predicted amplitudes in the 300-500ms timeframe for electrodes in the Central and Posterior ROIs. Error bars indicate standard error. A more negative amplitude represents a higher-amplitude N400. It should be noted that the y-axis is reversed, following convention.}
    \label{fig:predicted_values}
\end{figure}

Figure \ref{fig:predicted_values} shows both the true and predicted amplitudes in the 300-500ms time window for electrodes in the Central and Posterior ROIs (the canonical N400 ROIs). As can be seen, the N400 amplitudes predicted by GPT-2 surprisal are closer to the true amplitudes than those predicted by JRNN surprisal in 3 out of 4 of the conditions: \textsc{Best Completion} (One-tailed t-test testing whether N400 amplitude predicted by the GPT-2 surprisal model $<$ N400 amplitude predicted by the JRNN surprisal model: $t(10198) = 5.9324, p < 0.0001$), \textsc{Related} (GPT-2 $<$ JRNN: $t(9105.3) = 3.3548, p = 0.0019$), and \textsc{Implausible} (GPT-2 $>$ JRNN: $t(9363.7) = -7.4523, p < 0.0001$). While there is a difference in means in the expected direction for \textsc{Unrelated} completions, the difference is not significant (GPT-2 $>$ JRNN: $t(9695.5) = -1.2441, p = 0.4584$).

We further tested whether the models successfully predicted the differences between conditions by running one-tailed t-tests between the predicted amplitudes for conditions closest in mean predicted amplitude (using the values for the Central and Posterior ROIs, as shown in Figure \ref{fig:predicted_values}). GPT-2 surprisal successfully predicts that \textsc{Related} words elicit a higher-amplitude N400 than \textsc{Best Completion}s ($t(9601.2) = 18.466, p < 0.0001$), that \textsc{Unrelated} words elicit a higher-amplitude N400 than \textsc{Related} words ($t(9393.7) = 5.8936, p < 0.0001$), and \textsc{Implausible} words elicit a higher-amplitude N400 than \textsc{Unrelated} words ($t(9536.7) = 47.936, p < 0.0001$). On the other hand, while JRNN surprisal successfully predicts that \textsc{Related} words will elicit a higher-amplitude N400 than \textsc{Best Completions} ($t(9593.2) = 15.871, p < 0.0001$), and that \textsc{Implausible} words elicit a higher-amplitude N400 than \textsc{Unrelated} words ($t(9507.8) = 40.43, p < 0.0001$), it does not predict that \textsc{Unrelated} completions elicit a higher-amplitude N400 than \textsc{Related} completions ($t(9392.6) = 1.2691, p = 0.4584$), which was observed in the ERP data and was successfully predicted by GPT-2 surprisal. In fact, while its predictions are closer, in terms of predicting significant differences between conditions, JRNN surprisal does no better than cloze (Predicted \textsc{Best Completion} N400 amplitude $<$ predicted \textsc{Related} N400 amplitude: $t(9651.3) = 38.472, p < 0.0001$; \textsc{Related} $<$ \textsc{Unrelated}: $t(9384.9) = 0.5747, p = 1$; \textsc{Unrelated} $<$ \textsc{Implausible}: $t(9528.4) = 3.7758, p = 0.0004$).

\subsection{Discussion}
There are several key results from this study. First, surprisal and its interaction with scalp ROI are significant predictors of N400 amplitude, replicating previous results \cite{frank_2015_ERPResponseAmount,aurnhammer_2019_EvaluatingInformationtheoreticMeasures,merkx_2020_ComparingTransformersRNNs}. Second, we replicate the finding that the surprisal of T-LMs better predicts N400 amplitude than that of RNN-LMs \cite{merkx_2020_ComparingTransformersRNNs}. Finally, we find that while GPT-2 surprisal successfully predicts differences in N400 due to experimental manipulation---with \textsc{Best Completion} eliciting the most reduced N400, and \textsc{Related}, \textsc{Unrelated}, and \textsc{Implausible} completions each eliciting increasingly more negative (less positive) N400s---JRNN surprisal fails to distinguish between \textsc{Related} and \textsc{Unrelated} completions. This difficulty for the JRNN to predict the difference between the two is consistent with the findings of \citeA{michaelov_2020_HowWellDoes}.

Why does GPT-2 surprisal fit N400 amplitude better in both models? It is likely that there are multiple reasons for this, some of which may involve the simple fact of larger size and larger training data. It is also important to note that it is possible that the better predictions of GPT-2 surprisal are due to different factors across the experimental conditions.

However, one architecture-related possibility is suggested by our replication of the finding that JRNN surprisal can struggle to predict the difference between the \textsc{Related} and \textsc{Unelated} conditions for some sets of stimuli \cite{michaelov_2020_HowWellDoes} and our novel finding that this is not the case with GPT-2 surprisal. As can be seen in Figure \ref{fig:predicted_values}, the inability of JRNN surprisal to predict the difference between the two seems to be mostly driven by overestimating the N400 amplitude for the \textsc{Related} condition. This is something which is improved upon by using GPT-2 surprisal for prediction. It is important to note that as explained previously, the effect of semantic relatedness on N400 amplitude has been previously hypothesized to involve spreading activation or some other form of shallow semantic facilitation.

This is crucial because as \citeA{merkx_2020_ComparingTransformersRNNs} note, one of the key differences between T-LMs and RNN-LMs is that T-LMs do not have the same memory bottleneck as RNN-LMs---they have direct access to all previous words in the sequence. RNN-LMs, on the other hand, only have one current state, which is adjusted with each new input. Thus, this increased memory capacity---the capacity to `remember' exactly which words precede the current word---means that it is possible for the network to use specific previous words in predicting the next word, and that these could independently (or in a bag-of-words fashion) semantically facilitate predictions. This may also surface as susceptibility to priming---previous work has found that T-LMs are likely to repeat words that that they have already seen, and their predictions can be semantically primed by presenting them with an individual prime word \cite{misra_2020_ExploringBERTSensitivity}. 

\section{Experiment 2: Quantifying semantic facilitation}
In Experiment 2, we investigate whether the surprisal of GPT-2 incorporates something roughly analogous to spreading activation. That is, the finding that words are at least partly predicted because they have been activated by the semantics of previous words in the sequence. We do so by comparing the extent to which surprisal in each of the two models correlates with estimates of semantic similarity from each of the two models. A higher correlation would indicate the model is more biased towards predicting words that are semantically related to preceding words in the utterance, i.e., that it exhibits behavior akin to that commonly attributed to semantic spreading activation in humans.

\subsection{Method}
The high-level process for calculating semantic similarity in the NLMs was similar to that used for calculating surprisal. Stimuli from the same experiment \cite{bardolph_2018_SingleTrialEEG} were run through the NLMs, and the activation states of the model were recorded. For this study, however, we used the context-free word embeddings for each of the two NLMs (JRNN and GPT-2). As in previous work, we calculated the mean embeddings of all words preceding the target word, and calculated the cosine distance between this and the target word embedding. We then compared the extent to which cosine distance and surprisal were correlated for each network.

\subsection{Results and Discussion}

\begin{figure}
    \centering
    \includegraphics[width = \columnwidth]{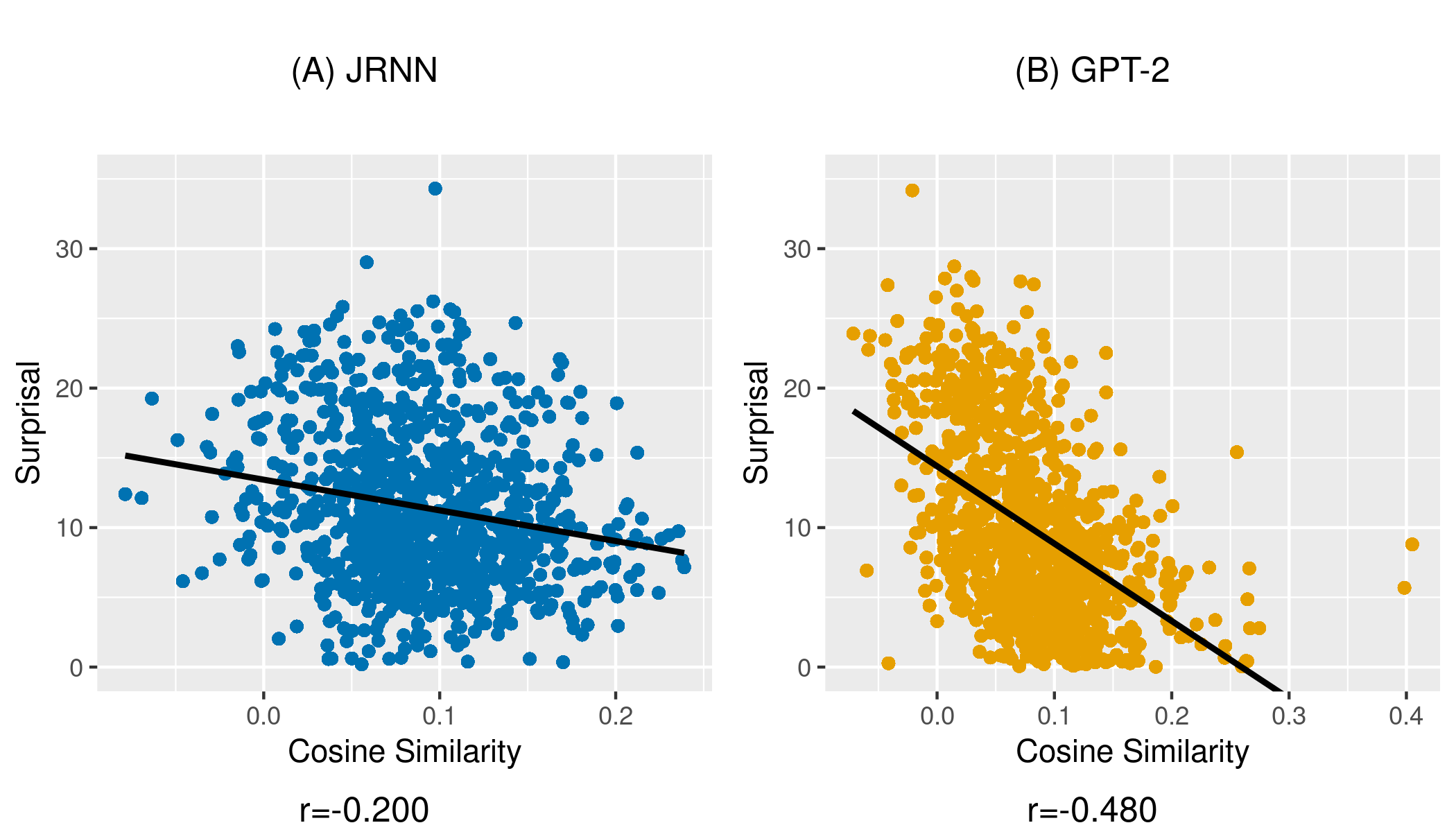}
    \caption{Cosine similarity and surprisal in the JRNN and GPT-2. $r$ is Pearson's correlation coefficient.}
    \label{fig:corplots}
\end{figure}

As can be seen in Figure \ref{fig:corplots}, cosine similarity and surprisal are substantially more correlated for GPT-2 ($r=-0.480$) than for the JRNN ($r=-0.200$), supporting the idea that semantic similarity is more directly correlated with surprisal in GPT-2 than the JRNN.

To the best of our knowledge the approach here is novel. Moreover, it is valuable because it is self-contained. We utilize each model's own semantic representations to evaluate the degree of semantic spreading activation. Since our own semantic representations are the only ones we have access to during language comprehension, the approach can be seen as cognitively plausible. Further, it can easily be applied to any language model and any data set, without necessarily requiring carefully constructed stimuli. 

\section{Experiment 3: Testing the implicit semantic facilitation account}
Experiment 1 showed that GPT-2 surprisal is a better predictor of N400 amplitude than JRNN surprisal is. Experiment 2 showed that GPT-2 surprisal is more highly correlated with semantic similarity than JRNN surprisal is. In Experiment 3, we directly test whether the latter finding explains the former, that is, whether the fact that GPT-2 surprisal is more correlated with semantic similarity (as operationalized by cosine similarity) leads to its better prediction of N400 amplitude.

\subsection{Method}
To do this, we investigated the extent to which each NLM's surprisal and cosine similarity metrics explain different proportions of the variance in N400 amplitude. If it is indeed the case that GPT-2 surprisal predicts N400 amplitude better than JRNN surprisal because its surprisal is more correlated with cosine similarity, then we should expect that adding cosine similarity as a predictor to a linear mixed-effects model with GPT-2 surprisal as a predictor should lead to less improvement than adding cosine similarity to the equivalent JRNN surprisal model.

\subsection{Results}
We tested this hypothesis by running likelihood ratio tests comparing the previous best models (including surprisal, ROI, and their interaction as fixed effects) with models that also included cosine similarity as a predictor and larger models that included both cosine similarity and its interaction with ROI as predictors. We found that each of these significantly improved the JRNN model fit (cosine similarity: $\chi^2(1) = 92.782, p < 0.0001$; cosine similarity x ROI: $\chi^2(6) = 93.594, p < 0.0001$), but neither improved GPT-2 model fit (cosine similarity: $\chi^2(1) = 0.1524, p = 1$; cosine similarity x ROI: $\chi^2(6) = 5.7346, p = 1$).
  
\subsection{Discussion}
The results show that cosine similarity explains additional variance in N400 amplitude beyond what is explained by JRNN surprisal. However, this is not the case with GPT-2 surprisal. The results, therefore, provide evidence that it is not only the case that GPT-2 surprisal is a better predictor of N400 amplitude and more correlated with semantic similarity than JRNN surprisal, but that the two are related. Specifically, they provide evidence for the hypothesis that it is the fact that GPT-2 surprisal correlates better with semantic similarity that makes it better at predicting N400 amplitude. 

\section{General Discussion}
Our results replicate and build upon previous work. As in previous work, we find that the surprisal of both RNN-LMs and T-LMs significantly predicts N400 amplitude, that T-LM surprisal is a better predictor than RNN-LM surprisal, and that JRNN surprisal struggles to predict the difference between words that are semantically related and unrelated to the highest-cloze completion \cite{frank_2015_ERPResponseAmount,aurnhammer_2019_EvaluatingInformationtheoreticMeasures,michaelov_2020_HowWellDoes,merkx_2020_ComparingTransformersRNNs}.

Our first novel finding is that GPT-2 surprisal can better predict the amplitude of N400s elicited by words that are semantically related to the highest-cloze completions than can JRNN surprisal. This allows GPT-2 surprisal to successfully distinguish between low-cloze words that are semantically related and unrelated to the best completion where JRNN surprisal struggles.

Our second and more important finding is that GPT-2 surprisal is more (inversely) correlated with GPT-2 semantic similarity than JRNN surprisal is with JRNN semantic similarity. We hypothesize that this is due to the difference in architecture---with access to the previous words in a given input, T-LMs are able to predict the next word based on any of these words, while RNN-LMs are limited to predict based on their single recurrent state, which may be multi-layered, but nonetheless does not store individual previous words explicitly. This is in concord with recent findings showing that BERT, another T-LM, is susceptible to priming \cite{ettinger_2020_WhatBERTNot,misra_2020_ExploringBERTSensitivity}. This suggests that `bag-of-words' semantic spreading activation, while not the whole story behind the neurocognitive system or systems underlying the N400 response, may still play a part in it, and thus, in the whole language comprehension system.

As a computational modeling study, the results of our experiments do not directly demonstrate a specific way in which language comprehension is implemented in the brain. However, they do demonstrate that it is not necessary to posit separate systems to explain the fact that N400 amplitude is affected both by how predictable an upcoming word is and the prior occurrence of semantically or associatively related words in the context. While this is been previously shown with other (more elaborate) modeling approaches (e.g. \citeNP{rabovsky_2018_ModellingN400Brain}), we show directly that lexical prediction could in principle implicitly incorporate semantic relatedness or similarity. 

It should be noted, however, that we do not provide evidence that the the same system \textit{must} underlie both kinds of N400 response. This is still an open research question. The one fMRI-based study on the topic, for example, suggests that the two may occur in distinct areas \cite{frank_2017_WordPredictabilitySemantic}. By contrast, the ERP-based work on the topic suggests that while there are differences between the effects of semantic facilitation and prediction that might be used to dissociate the two, the N400 response to each exhibit a very similar time course and topography \cite{kutas_1993_CompanyOtherWords,vanpetten_1993_ComparisonLexicalSentencelevel,lau_2013_DissociatingN400Effects,broderick_2018_ElectrophysiologicalCorrelatesSemantic}.

Another interesting question raised by our results is how to determine the cognitive plausibility of an NLM. As noted by \citeA{merkx_2020_ComparingTransformersRNNs}, intuitively, the RNN-LM architecture appears more cognitively plausible as a model of language comprehension than the T-LM. This is due to what \citeA{keller_2010_CognitivelyPlausibleModels} refers to as the distance-based memory cost of plausible language models---they have limited `working memory', and like humans, struggle with long-distance linguistic phenomena (e.g. long-range dependencies). This is something inherent in the architecture of RNN-LMs, even those with features such as long short-term memory (LSTM) that help them remember and forget necessary input. Transformers, on the other hand, have perfect memory for their entire context window (1024 tokens in GPT-2).

However, as discussed, some aspects of language comprehension may involve facilitation based on semantic similarity, and as we demonstrate in Experiment 2, this is a feature that appears to be more present in GPT-2, and combined with the findings of Ettinger and colleagues \cite{ettinger_2020_WhatBERTNot,misra_2020_ExploringBERTSensitivity}, the present study suggests that this may be a widespread feature of T-LMs in general. Therefore, it appears that some aspects of human language comprehension, specifically, those associated with language processing with limited working memory, may be better modeled with RNN-LM surprisal; while those that involve more shallow semantic facilitation may be better modeled with T-LMs surprisal. This may help to explain why there are conflicting results regarding which is better for modeling reading time \cite{merkx_2020_ComparingTransformersRNNs,wilcox_2020_PredictivePowerNeural,eisape_2020_ClozeDistillationImproves}.

Understanding the differences between the two model architectures and how these relate to different aspects of human language comprehension may thus not only help us improve our language models, but also offer insight into the neurocognitive systems involved in language.

\section{Acknowledgments}
This research was partially funded by a 2020-2021 CARTA (Center for Academic Research and Training in Anthropogeny) Fellowship awarded to James Michaelov.
\bibliographystyle{apacite}

\setlength{\bibleftmargin}{.125in}
\setlength{\bibindent}{-\bibleftmargin}

\bibliography{CogSci_Template.bib}

\begin{thebibliography}{}

\bibitem [\protect \citeauthoryear {%
Aurnhammer%
\ \BBA {} Frank%
}{%
Aurnhammer%
\ \BBA {} Frank%
}{%
{\protect \APACyear {2019}}%
}]{%
aurnhammer_2019_EvaluatingInformationtheoreticMeasures}
\APACinsertmetastar {%
aurnhammer_2019_EvaluatingInformationtheoreticMeasures}%
\begin{APACrefauthors}%
Aurnhammer, C.%
\BCBT {}\ \BBA {} Frank, S\BPBI L.%
\end{APACrefauthors}%
\unskip\
\newblock
\APACrefYearMonthDay{2019}{}{}.
\newblock
{\BBOQ}\APACrefatitle {Evaluating Information-Theoretic Measures of Word
  Prediction in Naturalistic Sentence Reading} {Evaluating
  information-theoretic measures of word prediction in naturalistic sentence
  reading}.{\BBCQ}
\newblock
\APACjournalVolNumPages{Neuropsychologia}{134}{}{107198}.
\PrintBackRefs{\CurrentBib}

\bibitem [\protect \citeauthoryear {%
Bardolph%
, Van~Petten%
\BCBL {}\ \BBA {} Coulson%
}{%
Bardolph%
\ \protect \BOthers {.}}{%
{\protect \APACyear {2018}}%
}]{%
bardolph_2018_SingleTrialEEG}
\APACinsertmetastar {%
bardolph_2018_SingleTrialEEG}%
\begin{APACrefauthors}%
Bardolph, M.%
, Van~Petten, C.%
\BCBL {}\ \BBA {} Coulson, S.%
\end{APACrefauthors}%
\unskip\
\newblock
\APACrefYearMonthDay{2018}{}{}.
\newblock
{\BBOQ}\APACrefatitle {Single {{Trial EEG Data Reveals Sensitivity}} to
  {{Conceptual Expectations}} ({{N400}}) and {{Integrative Demands}} ({{LPC}})}
  {Single {{Trial EEG Data Reveals Sensitivity}} to {{Conceptual Expectations}}
  ({{N400}}) and {{Integrative Demands}} ({{LPC}})}.{\BBCQ}
\newblock
\BIn{} \APACrefbtitle {Twelfth {{Annual Meeting}} of the {{Society}} for the
  {{Neurobiology}} of {{Language}}.} {Twelfth {{Annual Meeting}} of the
  {{Society}} for the {{Neurobiology}} of {{Language}}.}
\newblock
\APACaddressPublisher{{Quebec City, Canada}}{}.
\PrintBackRefs{\CurrentBib}

\bibitem [\protect \citeauthoryear {%
Benjamini%
\ \BBA {} Yekutieli%
}{%
Benjamini%
\ \BBA {} Yekutieli%
}{%
{\protect \APACyear {2001}}%
}]{%
benjamini2001control}
\APACinsertmetastar {%
benjamini2001control}%
\begin{APACrefauthors}%
Benjamini, Y.%
\BCBT {}\ \BBA {} Yekutieli, D.%
\end{APACrefauthors}%
\unskip\
\newblock
\APACrefYearMonthDay{2001}{}{}.
\newblock
{\BBOQ}\APACrefatitle {The control of the false discovery rate in multiple
  testing under dependency} {The control of the false discovery rate in
  multiple testing under dependency}.{\BBCQ}
\newblock
\APACjournalVolNumPages{Annals of statistics}{}{}{1165--1188}.
\PrintBackRefs{\CurrentBib}

\bibitem [\protect \citeauthoryear {%
Broderick%
, Anderson%
, Di~Liberto%
, Crosse%
\BCBL {}\ \BBA {} Lalor%
}{%
Broderick%
\ \protect \BOthers {.}}{%
{\protect \APACyear {2018}}%
}]{%
broderick_2018_ElectrophysiologicalCorrelatesSemantic}
\APACinsertmetastar {%
broderick_2018_ElectrophysiologicalCorrelatesSemantic}%
\begin{APACrefauthors}%
Broderick, M\BPBI P.%
, Anderson, A\BPBI J.%
, Di~Liberto, G\BPBI M.%
, Crosse, M\BPBI J.%
\BCBL {}\ \BBA {} Lalor, E\BPBI C.%
\end{APACrefauthors}%
\unskip\
\newblock
\APACrefYearMonthDay{2018}{}{}.
\newblock
{\BBOQ}\APACrefatitle {Electrophysiological {{Correlates}} of {{Semantic
  Dissimilarity Reflect}} the {{Comprehension}} of {{Natural}}, {{Narrative
  Speech}}} {Electrophysiological {{Correlates}} of {{Semantic Dissimilarity
  Reflect}} the {{Comprehension}} of {{Natural}}, {{Narrative Speech}}}.{\BBCQ}
\newblock
\APACjournalVolNumPages{Current Biology}{28}{5}{803-809.e3}.
\PrintBackRefs{\CurrentBib}

\bibitem [\protect \citeauthoryear {%
Brothers%
\ \BBA {} Kuperberg%
}{%
Brothers%
\ \BBA {} Kuperberg%
}{%
{\protect \APACyear {2021}}%
}]{%
brothers_2021_WordPredictabilityEffects}
\APACinsertmetastar {%
brothers_2021_WordPredictabilityEffects}%
\begin{APACrefauthors}%
Brothers, T.%
\BCBT {}\ \BBA {} Kuperberg, G\BPBI R.%
\end{APACrefauthors}%
\unskip\
\newblock
\APACrefYearMonthDay{2021}{{\APACmonth{02}}}{}.
\newblock
{\BBOQ}\APACrefatitle {Word Predictability Effects Are Linear, Not Logarithmic:
  {{Implications}} for Probabilistic Models of Sentence Comprehension} {Word
  predictability effects are linear, not logarithmic: {{Implications}} for
  probabilistic models of sentence comprehension}.{\BBCQ}
\newblock
\APACjournalVolNumPages{Journal of Memory and Language}{116}{}{104174}.
\newblock
\begin{APACrefDOI} \doi{10.1016/j.jml.2020.104174} \end{APACrefDOI}
\PrintBackRefs{\CurrentBib}

\bibitem [\protect \citeauthoryear {%
Brouwer%
, Fitz%
\BCBL {}\ \BBA {} Hoeks%
}{%
Brouwer%
\ \protect \BOthers {.}}{%
{\protect \APACyear {2012}}%
}]{%
brouwer_2012_GettingRealSemantica}
\APACinsertmetastar {%
brouwer_2012_GettingRealSemantica}%
\begin{APACrefauthors}%
Brouwer, H.%
, Fitz, H.%
\BCBL {}\ \BBA {} Hoeks, J.%
\end{APACrefauthors}%
\unskip\
\newblock
\APACrefYearMonthDay{2012}{}{}.
\newblock
{\BBOQ}\APACrefatitle {Getting Real about {{Semantic Illusions}}:
  {{Rethinking}} the Functional Role of the {{P600}} in Language Comprehension}
  {Getting real about {{Semantic Illusions}}: {{Rethinking}} the functional
  role of the {{P600}} in language comprehension}.{\BBCQ}
\newblock
\APACjournalVolNumPages{Brain Research}{1446}{}{127--143}.
\PrintBackRefs{\CurrentBib}

\bibitem [\protect \citeauthoryear {%
DeLong%
\ \BBA {} Kutas%
}{%
DeLong%
\ \BBA {} Kutas%
}{%
{\protect \APACyear {2020}}%
}]{%
delong_2020_ComprehendingSurprisingSentences}
\APACinsertmetastar {%
delong_2020_ComprehendingSurprisingSentences}%
\begin{APACrefauthors}%
DeLong, K\BPBI A.%
\BCBT {}\ \BBA {} Kutas, M.%
\end{APACrefauthors}%
\unskip\
\newblock
\APACrefYearMonthDay{2020}{}{}.
\newblock
{\BBOQ}\APACrefatitle {Comprehending Surprising Sentences: Sensitivity of
  Post-{{N400}} Positivities to Contextual Congruity and Semantic Relatedness}
  {Comprehending surprising sentences: Sensitivity of post-{{N400}}
  positivities to contextual congruity and semantic relatedness}.{\BBCQ}
\newblock
\APACjournalVolNumPages{Language, Cognition and Neuroscience}{0}{0}{1--20}.
\PrintBackRefs{\CurrentBib}

\bibitem [\protect \citeauthoryear {%
Eisape%
, Zaslavsky%
\BCBL {}\ \BBA {} Levy%
}{%
Eisape%
\ \protect \BOthers {.}}{%
{\protect \APACyear {2020}}%
}]{%
eisape_2020_ClozeDistillationImproves}
\APACinsertmetastar {%
eisape_2020_ClozeDistillationImproves}%
\begin{APACrefauthors}%
Eisape, T.%
, Zaslavsky, N.%
\BCBL {}\ \BBA {} Levy, R.%
\end{APACrefauthors}%
\unskip\
\newblock
\APACrefYearMonthDay{2020}{}{}.
\newblock
{\BBOQ}\APACrefatitle {Cloze {{Distillation Improves Psychometric Predictive
  Power}}} {Cloze {{Distillation Improves Psychometric Predictive
  Power}}}.{\BBCQ}
\newblock
\BIn{} \APACrefbtitle {Proceedings of the 24th {{Conference}} on
  {{Computational Natural Language Learning}}} {Proceedings of the 24th
  {{Conference}} on {{Computational Natural Language Learning}}}\ (\BPGS\
  609--619).
\newblock
\APACaddressPublisher{{Online}}{{Association for Computational Linguistics}}.
\PrintBackRefs{\CurrentBib}

\bibitem [\protect \citeauthoryear {%
Elman%
}{%
Elman%
}{%
{\protect \APACyear {1990}}%
}]{%
elman_1990_FindingStructureTimea}
\APACinsertmetastar {%
elman_1990_FindingStructureTimea}%
\begin{APACrefauthors}%
Elman, J\BPBI L.%
\end{APACrefauthors}%
\unskip\
\newblock
\APACrefYearMonthDay{1990}{}{}.
\newblock
{\BBOQ}\APACrefatitle {Finding {{Structure}} in {{Time}}} {Finding
  {{Structure}} in {{Time}}}.{\BBCQ}
\newblock
\APACjournalVolNumPages{Cognitive Science}{14}{2}{179--211}.
\PrintBackRefs{\CurrentBib}

\bibitem [\protect \citeauthoryear {%
Ettinger%
}{%
Ettinger%
}{%
{\protect \APACyear {2020}}%
}]{%
ettinger_2020_WhatBERTNot}
\APACinsertmetastar {%
ettinger_2020_WhatBERTNot}%
\begin{APACrefauthors}%
Ettinger, A.%
\end{APACrefauthors}%
\unskip\
\newblock
\APACrefYearMonthDay{2020}{}{}.
\newblock
{\BBOQ}\APACrefatitle {What {{BERT Is Not}}: {{Lessons}} from a {{New Suite}}
  of {{Psycholinguistic Diagnostics}} for {{Language Models}}} {What {{BERT Is
  Not}}: {{Lessons}} from a {{New Suite}} of {{Psycholinguistic Diagnostics}}
  for {{Language Models}}}.{\BBCQ}
\newblock
\APACjournalVolNumPages{Transactions of the Association for Computational
  Linguistics}{8}{}{34--48}.
\PrintBackRefs{\CurrentBib}

\bibitem [\protect \citeauthoryear {%
Ettinger%
, Feldman%
, Resnik%
\BCBL {}\ \BBA {} Phillips%
}{%
Ettinger%
\ \protect \BOthers {.}}{%
{\protect \APACyear {2016}}%
}]{%
ettinger_2016_ModelingN400Amplitude}
\APACinsertmetastar {%
ettinger_2016_ModelingN400Amplitude}%
\begin{APACrefauthors}%
Ettinger, A.%
, Feldman, N.%
, Resnik, P.%
\BCBL {}\ \BBA {} Phillips, C.%
\end{APACrefauthors}%
\unskip\
\newblock
\APACrefYearMonthDay{2016}{}{}.
\newblock
{\BBOQ}\APACrefatitle {Modeling {{N400}} Amplitude Using Vector Space Models of
  Word Representation.} {Modeling {{N400}} amplitude using vector space models
  of word representation.}{\BBCQ}
\newblock
\BIn{} \APACrefbtitle {Proceedings of the 38th {{Annual Conference}} of the
  {{Cognitive Science Society}}.} {Proceedings of the 38th {{Annual
  Conference}} of the {{Cognitive Science Society}}.}
\newblock
\APACaddressPublisher{{Philadelphia, USA}}{}.
\PrintBackRefs{\CurrentBib}

\bibitem [\protect \citeauthoryear {%
Federmeier%
\ \BBA {} Kutas%
}{%
Federmeier%
\ \BBA {} Kutas%
}{%
{\protect \APACyear {1999}}%
}]{%
federmeier_1999_RoseAnyOther}
\APACinsertmetastar {%
federmeier_1999_RoseAnyOther}%
\begin{APACrefauthors}%
Federmeier, K\BPBI D.%
\BCBT {}\ \BBA {} Kutas, M.%
\end{APACrefauthors}%
\unskip\
\newblock
\APACrefYearMonthDay{1999}{}{}.
\newblock
{\BBOQ}\APACrefatitle {A {{Rose}} by {{Any Other Name}}: {{Long}}-{{Term Memory
  Structure}} and {{Sentence Processing}}} {A {{Rose}} by {{Any Other Name}}:
  {{Long}}-{{Term Memory Structure}} and {{Sentence Processing}}}.{\BBCQ}
\newblock
\APACjournalVolNumPages{Journal of Memory and Language}{41}{4}{469--495}.
\PrintBackRefs{\CurrentBib}

\bibitem [\protect \citeauthoryear {%
Frank%
, Otten%
, Galli%
\BCBL {}\ \BBA {} Vigliocco%
}{%
Frank%
\ \protect \BOthers {.}}{%
{\protect \APACyear {2015}}%
}]{%
frank_2015_ERPResponseAmount}
\APACinsertmetastar {%
frank_2015_ERPResponseAmount}%
\begin{APACrefauthors}%
Frank, S\BPBI L.%
, Otten, L\BPBI J.%
, Galli, G.%
\BCBL {}\ \BBA {} Vigliocco, G.%
\end{APACrefauthors}%
\unskip\
\newblock
\APACrefYearMonthDay{2015}{}{}.
\newblock
{\BBOQ}\APACrefatitle {The {{ERP}} Response to the Amount of Information
  Conveyed by Words in Sentences} {The {{ERP}} response to the amount of
  information conveyed by words in sentences}.{\BBCQ}
\newblock
\APACjournalVolNumPages{Brain and Language}{140}{}{1--11}.
\PrintBackRefs{\CurrentBib}

\bibitem [\protect \citeauthoryear {%
Frank%
\ \BBA {} Willems%
}{%
Frank%
\ \BBA {} Willems%
}{%
{\protect \APACyear {2017}}%
}]{%
frank_2017_WordPredictabilitySemantic}
\APACinsertmetastar {%
frank_2017_WordPredictabilitySemantic}%
\begin{APACrefauthors}%
Frank, S\BPBI L.%
\BCBT {}\ \BBA {} Willems, R\BPBI M.%
\end{APACrefauthors}%
\unskip\
\newblock
\APACrefYearMonthDay{2017}{}{}.
\newblock
{\BBOQ}\APACrefatitle {Word Predictability and Semantic Similarity Show
  Distinct Patterns of Brain Activity during Language Comprehension} {Word
  predictability and semantic similarity show distinct patterns of brain
  activity during language comprehension}.{\BBCQ}
\newblock
\APACjournalVolNumPages{Language, Cognition and
  Neuroscience}{32}{9}{1192--1203}.
\PrintBackRefs{\CurrentBib}

\bibitem [\protect \citeauthoryear {%
Jozefowicz%
, Vinyals%
, Schuster%
, Shazeer%
\BCBL {}\ \BBA {} Wu%
}{%
Jozefowicz%
\ \protect \BOthers {.}}{%
{\protect \APACyear {2016}}%
}]{%
jozefowicz_2016_ExploringLimitsLanguage}
\APACinsertmetastar {%
jozefowicz_2016_ExploringLimitsLanguage}%
\begin{APACrefauthors}%
Jozefowicz, R.%
, Vinyals, O.%
, Schuster, M.%
, Shazeer, N.%
\BCBL {}\ \BBA {} Wu, Y.%
\end{APACrefauthors}%
\unskip\
\newblock
\APACrefYearMonthDay{2016}{}{}.
\newblock
{\BBOQ}\APACrefatitle {Exploring the {{Limits}} of {{Language Modeling}}}
  {Exploring the {{Limits}} of {{Language Modeling}}}.{\BBCQ}
\newblock
\APACjournalVolNumPages{arXiv:1602.02410 [cs]}{}{}{}.
\PrintBackRefs{\CurrentBib}

\bibitem [\protect \citeauthoryear {%
Keller%
}{%
Keller%
}{%
{\protect \APACyear {2010}}%
}]{%
keller_2010_CognitivelyPlausibleModels}
\APACinsertmetastar {%
keller_2010_CognitivelyPlausibleModels}%
\begin{APACrefauthors}%
Keller, F.%
\end{APACrefauthors}%
\unskip\
\newblock
\APACrefYearMonthDay{2010}{{\APACmonth{07}}}{}.
\newblock
{\BBOQ}\APACrefatitle {Cognitively {{Plausible Models}} of {{Human Language
  Processing}}} {Cognitively {{Plausible Models}} of {{Human Language
  Processing}}}.{\BBCQ}
\newblock
\BIn{} \APACrefbtitle {Proceedings of the {{ACL}} 2010 {{Conference Short
  Papers}}} {Proceedings of the {{ACL}} 2010 {{Conference Short Papers}}}\
  (\BPGS\ 60--67).
\newblock
\APACaddressPublisher{{Uppsala, Sweden}}{{Association for Computational
  Linguistics}}.
\PrintBackRefs{\CurrentBib}

\bibitem [\protect \citeauthoryear {%
Kuperberg%
}{%
Kuperberg%
}{%
{\protect \APACyear {2016}}%
}]{%
kuperberg_2016_SeparateStreamsProbabilistic}
\APACinsertmetastar {%
kuperberg_2016_SeparateStreamsProbabilistic}%
\begin{APACrefauthors}%
Kuperberg, G\BPBI R.%
\end{APACrefauthors}%
\unskip\
\newblock
\APACrefYearMonthDay{2016}{}{}.
\newblock
{\BBOQ}\APACrefatitle {Separate Streams or Probabilistic Inference? {{What}}
  the {{N400}} Can Tell Us about the Comprehension of Events} {Separate streams
  or probabilistic inference? {{What}} the {{N400}} can tell us about the
  comprehension of events}.{\BBCQ}
\newblock
\APACjournalVolNumPages{Language, Cognition and Neuroscience}{31}{5}{602--616}.
\PrintBackRefs{\CurrentBib}

\bibitem [\protect \citeauthoryear {%
Kuperberg%
, Brothers%
\BCBL {}\ \BBA {} Wlotko%
}{%
Kuperberg%
\ \protect \BOthers {.}}{%
{\protect \APACyear {2020}}%
}]{%
kuperberg_2020_TaleTwoPositivities}
\APACinsertmetastar {%
kuperberg_2020_TaleTwoPositivities}%
\begin{APACrefauthors}%
Kuperberg, G\BPBI R.%
, Brothers, T.%
\BCBL {}\ \BBA {} Wlotko, E\BPBI W.%
\end{APACrefauthors}%
\unskip\
\newblock
\APACrefYearMonthDay{2020}{}{}.
\newblock
{\BBOQ}\APACrefatitle {A {{Tale}} of {{Two Positivities}} and the {{N400}}:
  {{Distinct Neural Signatures Are Evoked}} by {{Confirmed}} and {{Violated
  Predictions}} at {{Different Levels}} of {{Representation}}} {A {{Tale}} of
  {{Two Positivities}} and the {{N400}}: {{Distinct Neural Signatures Are
  Evoked}} by {{Confirmed}} and {{Violated Predictions}} at {{Different
  Levels}} of {{Representation}}}.{\BBCQ}
\newblock
\APACjournalVolNumPages{Journal of Cognitive Neuroscience}{32}{1}{12--35}.
\PrintBackRefs{\CurrentBib}

\bibitem [\protect \citeauthoryear {%
Kuperberg%
, Sitnikova%
, Caplan%
\BCBL {}\ \BBA {} Holcomb%
}{%
Kuperberg%
\ \protect \BOthers {.}}{%
{\protect \APACyear {2003}}%
}]{%
kuperberg_2003_ElectrophysiologicalDistinctionsProcessing}
\APACinsertmetastar {%
kuperberg_2003_ElectrophysiologicalDistinctionsProcessing}%
\begin{APACrefauthors}%
Kuperberg, G\BPBI R.%
, Sitnikova, T.%
, Caplan, D.%
\BCBL {}\ \BBA {} Holcomb, P\BPBI J.%
\end{APACrefauthors}%
\unskip\
\newblock
\APACrefYearMonthDay{2003}{{\APACmonth{06}}}{}.
\newblock
{\BBOQ}\APACrefatitle {Electrophysiological Distinctions in Processing
  Conceptual Relationships within Simple Sentences} {Electrophysiological
  distinctions in processing conceptual relationships within simple
  sentences}.{\BBCQ}
\newblock
\APACjournalVolNumPages{Cognitive Brain Research}{17}{1}{117--129}.
\newblock
\begin{APACrefDOI} \doi{10.1016/S0926-6410(03)00086-7} \end{APACrefDOI}
\PrintBackRefs{\CurrentBib}

\bibitem [\protect \citeauthoryear {%
Kutas%
}{%
Kutas%
}{%
{\protect \APACyear {1993}}%
}]{%
kutas_1993_CompanyOtherWords}
\APACinsertmetastar {%
kutas_1993_CompanyOtherWords}%
\begin{APACrefauthors}%
Kutas, M.%
\end{APACrefauthors}%
\unskip\
\newblock
\APACrefYearMonthDay{1993}{}{}.
\newblock
{\BBOQ}\APACrefatitle {In the Company of Other Words: {{Electrophysiological}}
  Evidence for Single-Word and Sentence Context Effects} {In the company of
  other words: {{Electrophysiological}} evidence for single-word and sentence
  context effects}.{\BBCQ}
\newblock
\APACjournalVolNumPages{Language and Cognitive Processes}{8}{4}{533--572}.
\PrintBackRefs{\CurrentBib}

\bibitem [\protect \citeauthoryear {%
Kutas%
\ \BBA {} Federmeier%
}{%
Kutas%
\ \BBA {} Federmeier%
}{%
{\protect \APACyear {2011}}%
}]{%
kutas_2011_ThirtyYearsCounting}
\APACinsertmetastar {%
kutas_2011_ThirtyYearsCounting}%
\begin{APACrefauthors}%
Kutas, M.%
\BCBT {}\ \BBA {} Federmeier, K\BPBI D.%
\end{APACrefauthors}%
\unskip\
\newblock
\APACrefYearMonthDay{2011}{}{}.
\newblock
{\BBOQ}\APACrefatitle {Thirty {{Years}} and {{Counting}}: {{Finding Meaning}}
  in the {{N400 Component}} of the {{Event}}-{{Related Brain Potential}}
  ({{ERP}})} {Thirty {{Years}} and {{Counting}}: {{Finding Meaning}} in the
  {{N400 Component}} of the {{Event}}-{{Related Brain Potential}}
  ({{ERP}})}.{\BBCQ}
\newblock
\APACjournalVolNumPages{Annual Review of Psychology}{62}{1}{621--647}.
\PrintBackRefs{\CurrentBib}

\bibitem [\protect \citeauthoryear {%
Kutas%
\ \BBA {} Hillyard%
}{%
Kutas%
\ \BBA {} Hillyard%
}{%
{\protect \APACyear {1980}}%
}]{%
kutas_1980_ReadingSenselessSentences}
\APACinsertmetastar {%
kutas_1980_ReadingSenselessSentences}%
\begin{APACrefauthors}%
Kutas, M.%
\BCBT {}\ \BBA {} Hillyard, S\BPBI A.%
\end{APACrefauthors}%
\unskip\
\newblock
\APACrefYearMonthDay{1980}{}{}.
\newblock
{\BBOQ}\APACrefatitle {Reading Senseless Sentences: Brain Potentials Reflect
  Semantic Incongruity} {Reading senseless sentences: Brain potentials reflect
  semantic incongruity}.{\BBCQ}
\newblock
\APACjournalVolNumPages{Science}{207}{4427}{203--205}.
\PrintBackRefs{\CurrentBib}

\bibitem [\protect \citeauthoryear {%
Kutas%
\ \BBA {} Van~Petten%
}{%
Kutas%
\ \BBA {} Van~Petten%
}{%
{\protect \APACyear {1988}}%
}]{%
kutas_1988_EventrelatedBrainPotential}
\APACinsertmetastar {%
kutas_1988_EventrelatedBrainPotential}%
\begin{APACrefauthors}%
Kutas, M.%
\BCBT {}\ \BBA {} Van~Petten, C.%
\end{APACrefauthors}%
\unskip\
\newblock
\APACrefYearMonthDay{1988}{}{}.
\newblock
{\BBOQ}\APACrefatitle {Event-Related Brain Potential Studies of Language}
  {Event-related brain potential studies of language}.{\BBCQ}
\newblock
\APACjournalVolNumPages{Advances in psychophysiology}{3}{}{139--187}.
\PrintBackRefs{\CurrentBib}

\bibitem [\protect \citeauthoryear {%
Lau%
, Holcomb%
\BCBL {}\ \BBA {} Kuperberg%
}{%
Lau%
\ \protect \BOthers {.}}{%
{\protect \APACyear {2013}}%
}]{%
lau_2013_DissociatingN400Effects}
\APACinsertmetastar {%
lau_2013_DissociatingN400Effects}%
\begin{APACrefauthors}%
Lau, E\BPBI F.%
, Holcomb, P\BPBI J.%
\BCBL {}\ \BBA {} Kuperberg, G\BPBI R.%
\end{APACrefauthors}%
\unskip\
\newblock
\APACrefYearMonthDay{2013}{}{}.
\newblock
{\BBOQ}\APACrefatitle {Dissociating {{N400 Effects}} of {{Prediction}} from
  {{Association}} in {{Single}}-Word {{Contexts}}} {Dissociating {{N400
  Effects}} of {{Prediction}} from {{Association}} in {{Single}}-word
  {{Contexts}}}.{\BBCQ}
\newblock
\APACjournalVolNumPages{Journal of Cognitive Neuroscience}{25}{3}{484--502}.
\PrintBackRefs{\CurrentBib}

\bibitem [\protect \citeauthoryear {%
Luke%
\ \BBA {} Christianson%
}{%
Luke%
\ \BBA {} Christianson%
}{%
{\protect \APACyear {2016}}%
}]{%
luke_2016_LimitsLexicalPrediction}
\APACinsertmetastar {%
luke_2016_LimitsLexicalPrediction}%
\begin{APACrefauthors}%
Luke, S\BPBI G.%
\BCBT {}\ \BBA {} Christianson, K.%
\end{APACrefauthors}%
\unskip\
\newblock
\APACrefYearMonthDay{2016}{}{}.
\newblock
{\BBOQ}\APACrefatitle {Limits on Lexical Prediction during Reading} {Limits on
  lexical prediction during reading}.{\BBCQ}
\newblock
\APACjournalVolNumPages{Cognitive Psychology}{88}{}{22--60}.
\PrintBackRefs{\CurrentBib}

\bibitem [\protect \citeauthoryear {%
Merkx%
\ \BBA {} Frank%
}{%
Merkx%
\ \BBA {} Frank%
}{%
{\protect \APACyear {2020}}%
}]{%
merkx_2020_ComparingTransformersRNNs}
\APACinsertmetastar {%
merkx_2020_ComparingTransformersRNNs}%
\begin{APACrefauthors}%
Merkx, D.%
\BCBT {}\ \BBA {} Frank, S\BPBI L.%
\end{APACrefauthors}%
\unskip\
\newblock
\APACrefYearMonthDay{2020}{}{}.
\newblock
{\BBOQ}\APACrefatitle {Comparing {{Transformers}} and {{RNNs}} on Predicting
  Human Sentence Processing Data} {Comparing {{Transformers}} and {{RNNs}} on
  predicting human sentence processing data}.{\BBCQ}
\newblock
\APACjournalVolNumPages{arXiv:2005.09471 [cs]}{}{}{}.
\PrintBackRefs{\CurrentBib}

\bibitem [\protect \citeauthoryear {%
Michaelov%
\ \BBA {} Bergen%
}{%
Michaelov%
\ \BBA {} Bergen%
}{%
{\protect \APACyear {2020}}%
}]{%
michaelov_2020_HowWellDoes}
\APACinsertmetastar {%
michaelov_2020_HowWellDoes}%
\begin{APACrefauthors}%
Michaelov, J\BPBI A.%
\BCBT {}\ \BBA {} Bergen, B\BPBI K.%
\end{APACrefauthors}%
\unskip\
\newblock
\APACrefYearMonthDay{2020}{{\APACmonth{11}}}{}.
\newblock
{\BBOQ}\APACrefatitle {How Well Does Surprisal Explain {{N400}} Amplitude under
  Different Experimental Conditions?} {How well does surprisal explain {{N400}}
  amplitude under different experimental conditions?}{\BBCQ}
\newblock
\BIn{} \APACrefbtitle {Proceedings of the 24th {{Conference}} on
  {{Computational Natural Language Learning}} ({{CoNLL}} 2020).} {Proceedings
  of the 24th {{Conference}} on {{Computational Natural Language Learning}}
  ({{CoNLL}} 2020).}
\newblock
\APACaddressPublisher{Online}{{Association for Computational Linguistics}}.
\PrintBackRefs{\CurrentBib}

\bibitem [\protect \citeauthoryear {%
Misra%
, Ettinger%
\BCBL {}\ \BBA {} Rayz%
}{%
Misra%
\ \protect \BOthers {.}}{%
{\protect \APACyear {2020}}%
}]{%
misra_2020_ExploringBERTSensitivity}
\APACinsertmetastar {%
misra_2020_ExploringBERTSensitivity}%
\begin{APACrefauthors}%
Misra, K.%
, Ettinger, A.%
\BCBL {}\ \BBA {} Rayz, J.%
\end{APACrefauthors}%
\unskip\
\newblock
\APACrefYearMonthDay{2020}{}{}.
\newblock
{\BBOQ}\APACrefatitle {Exploring {{BERT}}'s {{Sensitivity}} to {{Lexical Cues}}
  Using {{Tests}} from {{Semantic Priming}}} {Exploring {{BERT}}'s
  {{Sensitivity}} to {{Lexical Cues}} using {{Tests}} from {{Semantic
  Priming}}}.{\BBCQ}
\newblock
\BIn{} \APACrefbtitle {Findings of the {{Association}} for {{Computational
  Linguistics}}: {{EMNLP}} 2020} {Findings of the {{Association}} for
  {{Computational Linguistics}}: {{EMNLP}} 2020}\ (\BPGS\ 4625--4635).
\newblock
\APACaddressPublisher{{Online}}{{Association for Computational Linguistics}}.
\PrintBackRefs{\CurrentBib}

\bibitem [\protect \citeauthoryear {%
Nieuwland%
\ \BBA {} Kuperberg%
}{%
Nieuwland%
\ \BBA {} Kuperberg%
}{%
{\protect \APACyear {2008}}%
}]{%
nieuwland2008truth}
\APACinsertmetastar {%
nieuwland2008truth}%
\begin{APACrefauthors}%
Nieuwland, M\BPBI S.%
\BCBT {}\ \BBA {} Kuperberg, G\BPBI R.%
\end{APACrefauthors}%
\unskip\
\newblock
\APACrefYearMonthDay{2008}{}{}.
\newblock
{\BBOQ}\APACrefatitle {When the truth is not too hard to handle: An
  event-related potential study on the pragmatics of negation} {When the truth
  is not too hard to handle: An event-related potential study on the pragmatics
  of negation}.{\BBCQ}
\newblock
\APACjournalVolNumPages{Psychological Science}{19}{12}{1213--1218}.
\PrintBackRefs{\CurrentBib}

\bibitem [\protect \citeauthoryear {%
Parviz%
, Johnson%
, Johnson%
\BCBL {}\ \BBA {} Brock%
}{%
Parviz%
\ \protect \BOthers {.}}{%
{\protect \APACyear {2011}}%
}]{%
parviz_2011_UsingLanguageModels}
\APACinsertmetastar {%
parviz_2011_UsingLanguageModels}%
\begin{APACrefauthors}%
Parviz, M.%
, Johnson, M.%
, Johnson, B.%
\BCBL {}\ \BBA {} Brock, J.%
\end{APACrefauthors}%
\unskip\
\newblock
\APACrefYearMonthDay{2011}{{\APACmonth{12}}}{}.
\newblock
{\BBOQ}\APACrefatitle {Using {{Language Models}} and {{Latent Semantic
  Analysis}} to {{Characterise}} the {{N400m Neural Response}}} {Using
  {{Language Models}} and {{Latent Semantic Analysis}} to {{Characterise}} the
  {{N400m Neural Response}}}.{\BBCQ}
\newblock
\BIn{} \APACrefbtitle {Proceedings of the {{Australasian Language Technology
  Association Workshop}} 2011} {Proceedings of the {{Australasian Language
  Technology Association Workshop}} 2011}\ (\BPGS\ 38--46).
\newblock
\APACaddressPublisher{{Canberra, Australia}}{}.
\PrintBackRefs{\CurrentBib}

\bibitem [\protect \citeauthoryear {%
{R Core Team}%
}{%
{R Core Team}%
}{%
{\protect \APACyear {2020}}%
}]{%
R_language}
\APACinsertmetastar {%
R_language}%
\begin{APACrefauthors}%
{R Core Team}.%
\end{APACrefauthors}%
\unskip\
\newblock
\APACrefYearMonthDay{2020}{}{}.
\newblock
{\BBOQ}\APACrefatitle {R: A Language and Environment for Statistical Computing}
  {R: A language and environment for statistical computing}{\BBCQ}\
  [\bibcomputersoftwaremanual].
\newblock
\APACaddressPublisher{Vienna, Austria}{}.
\newblock
\begin{APACrefURL} \url{https://www.R-project.org/} \end{APACrefURL}
\PrintBackRefs{\CurrentBib}

\bibitem [\protect \citeauthoryear {%
Rabovsky%
, Hansen%
\BCBL {}\ \BBA {} McClelland%
}{%
Rabovsky%
\ \protect \BOthers {.}}{%
{\protect \APACyear {2018}}%
}]{%
rabovsky_2018_ModellingN400Brain}
\APACinsertmetastar {%
rabovsky_2018_ModellingN400Brain}%
\begin{APACrefauthors}%
Rabovsky, M.%
, Hansen, S\BPBI S.%
\BCBL {}\ \BBA {} McClelland, J\BPBI L.%
\end{APACrefauthors}%
\unskip\
\newblock
\APACrefYearMonthDay{2018}{{\APACmonth{09}}}{}.
\newblock
{\BBOQ}\APACrefatitle {Modelling the {{N400}} Brain Potential as Change in a
  Probabilistic Representation of Meaning} {Modelling the {{N400}} brain
  potential as change in a probabilistic representation of meaning}.{\BBCQ}
\newblock
\APACjournalVolNumPages{Nature Human Behaviour}{2}{9}{693--705}.
\newblock
\begin{APACrefDOI} \doi{10.1038/s41562-018-0406-4} \end{APACrefDOI}
\PrintBackRefs{\CurrentBib}

\bibitem [\protect \citeauthoryear {%
Radford%
\ \protect \BOthers {.}}{%
Radford%
\ \protect \BOthers {.}}{%
{\protect \APACyear {2019}}%
}]{%
radford_2019_LanguageModelsAre}
\APACinsertmetastar {%
radford_2019_LanguageModelsAre}%
\begin{APACrefauthors}%
Radford, A.%
, Wu, J.%
, Child, R.%
, Luan, D.%
, Amodei, D.%
\BCBL {}\ \BBA {} Sutskever, I.%
\end{APACrefauthors}%
\unskip\
\newblock
\APACrefYearMonthDay{2019}{}{}.
\newblock
{\BBOQ}\APACrefatitle {Language {{Models}} Are {{Unsupervised Multitask
  Learners}}} {Language {{Models}} are {{Unsupervised Multitask
  Learners}}}.{\BBCQ}
\newblock
\APACjournalVolNumPages{}{}{}{24}.
\PrintBackRefs{\CurrentBib}

\bibitem [\protect \citeauthoryear {%
Taylor%
}{%
Taylor%
}{%
{\protect \APACyear {1957}}%
}]{%
taylor_1957_ClozeReadabilityScores}
\APACinsertmetastar {%
taylor_1957_ClozeReadabilityScores}%
\begin{APACrefauthors}%
Taylor, W\BPBI L.%
\end{APACrefauthors}%
\unskip\
\newblock
\APACrefYearMonthDay{1957}{}{}.
\newblock
{\BBOQ}\APACrefatitle {``{{Cloze}}'' Readability Scores as Indices of
  Individual Differences in Comprehension and Aptitude} {``{{Cloze}}''
  readability scores as indices of individual differences in comprehension and
  aptitude}.{\BBCQ}
\newblock
\APACjournalVolNumPages{Journal of Applied Psychology}{41}{1}{19--26}.
\PrintBackRefs{\CurrentBib}

\bibitem [\protect \citeauthoryear {%
Thornhill%
\ \BBA {} Van~Petten%
}{%
Thornhill%
\ \BBA {} Van~Petten%
}{%
{\protect \APACyear {2012}}%
}]{%
thornhill_2012_LexicalConceptualAnticipation}
\APACinsertmetastar {%
thornhill_2012_LexicalConceptualAnticipation}%
\begin{APACrefauthors}%
Thornhill, D\BPBI E.%
\BCBT {}\ \BBA {} Van~Petten, C.%
\end{APACrefauthors}%
\unskip\
\newblock
\APACrefYearMonthDay{2012}{}{}.
\newblock
{\BBOQ}\APACrefatitle {Lexical versus Conceptual Anticipation during Sentence
  Processing: {{Frontal}} Positivity and {{N400 ERP}} Components} {Lexical
  versus conceptual anticipation during sentence processing: {{Frontal}}
  positivity and {{N400 ERP}} components}.{\BBCQ}
\newblock
\APACjournalVolNumPages{International Journal of
  Psychophysiology}{83}{3}{382--392}.
\PrintBackRefs{\CurrentBib}

\bibitem [\protect \citeauthoryear {%
Van~Petten%
}{%
Van~Petten%
}{%
{\protect \APACyear {1993}}%
}]{%
vanpetten_1993_ComparisonLexicalSentencelevel}
\APACinsertmetastar {%
vanpetten_1993_ComparisonLexicalSentencelevel}%
\begin{APACrefauthors}%
Van~Petten, C.%
\end{APACrefauthors}%
\unskip\
\newblock
\APACrefYearMonthDay{1993}{}{}.
\newblock
{\BBOQ}\APACrefatitle {A Comparison of Lexical and Sentence-Level Context
  Effects in Event-Related Potentials} {A comparison of lexical and
  sentence-level context effects in event-related potentials}.{\BBCQ}
\newblock
\APACjournalVolNumPages{Language and Cognitive Processes}{8}{4}{485--531}.
\PrintBackRefs{\CurrentBib}

\bibitem [\protect \citeauthoryear {%
Van~Petten%
\ \BBA {} Luka%
}{%
Van~Petten%
\ \BBA {} Luka%
}{%
{\protect \APACyear {2012}}%
}]{%
vanpetten_2012_PredictionLanguageComprehension}
\APACinsertmetastar {%
vanpetten_2012_PredictionLanguageComprehension}%
\begin{APACrefauthors}%
Van~Petten, C.%
\BCBT {}\ \BBA {} Luka, B\BPBI J.%
\end{APACrefauthors}%
\unskip\
\newblock
\APACrefYearMonthDay{2012}{{\APACmonth{02}}}{}.
\newblock
{\BBOQ}\APACrefatitle {Prediction during Language Comprehension: {{Benefits}},
  Costs, and {{ERP}} Components} {Prediction during language comprehension:
  {{Benefits}}, costs, and {{ERP}} components}.{\BBCQ}
\newblock
\APACjournalVolNumPages{International Journal of
  Psychophysiology}{83}{2}{176--190}.
\newblock
\begin{APACrefDOI} \doi{10.1016/j.ijpsycho.2011.09.015} \end{APACrefDOI}
\PrintBackRefs{\CurrentBib}

\bibitem [\protect \citeauthoryear {%
Vaswani%
\ \protect \BOthers {.}}{%
Vaswani%
\ \protect \BOthers {.}}{%
{\protect \APACyear {2017}}%
}]{%
vaswani_2017_attention}
\APACinsertmetastar {%
vaswani_2017_attention}%
\begin{APACrefauthors}%
Vaswani, A.%
, Shazeer, N.%
, Parmar, N.%
, Uszkoreit, J.%
, Jones, L.%
, Gomez, A\BPBI N.%
\BDBL {}Polosukhin, I.%
\end{APACrefauthors}%
\unskip\
\newblock
\APACrefYearMonthDay{2017}{}{}.
\newblock
{\BBOQ}\APACrefatitle {Attention is All you Need} {Attention is all you
  need}.{\BBCQ}
\newblock
\BIn{} I.~Guyon\ \BOthers {.}\ (\BEDS), \APACrefbtitle {Advances in Neural
  Information Processing Systems} {Advances in neural information processing
  systems}\ (\BVOL~30, \BPGS\ 5998--6008).
\newblock
\APACaddressPublisher{}{Curran Associates, Inc.}
\PrintBackRefs{\CurrentBib}

\bibitem [\protect \citeauthoryear {%
Wilcox%
, Gauthier%
, Hu%
, Qian%
\BCBL {}\ \BBA {} Levy%
}{%
Wilcox%
\ \protect \BOthers {.}}{%
{\protect \APACyear {2020}}%
}]{%
wilcox_2020_PredictivePowerNeural}
\APACinsertmetastar {%
wilcox_2020_PredictivePowerNeural}%
\begin{APACrefauthors}%
Wilcox, E\BPBI G.%
, Gauthier, J.%
, Hu, J.%
, Qian, P.%
\BCBL {}\ \BBA {} Levy, R\BPBI P.%
\end{APACrefauthors}%
\unskip\
\newblock
\APACrefYearMonthDay{2020}{}{}.
\newblock
{\BBOQ}\APACrefatitle {On the {{Predictive Power}} of {{Neural Language
  Models}} for {{Human Real}}-{{Time Comprehension Behavior}}} {On the
  {{Predictive Power}} of {{Neural Language Models}} for {{Human Real}}-{{Time
  Comprehension Behavior}}}.{\BBCQ}
\newblock
\BIn{} \APACrefbtitle {Proceedings of the 42nd {{Annual Meeting}} of the
  {{Cognitive Science Society}} ({{CogSci}} 2020)} {Proceedings of the 42nd
  {{Annual Meeting}} of the {{Cognitive Science Society}} ({{CogSci}} 2020)}\
  (\BPG~7).
\PrintBackRefs{\CurrentBib}

\end{thebibliography}

\end{document}